\documentclass[runningheads]{llncs}
\usepackage{graphbox}

\usepackage{tikz}
\usepackage{comment}
\usepackage{amsmath,amssymb} 
\usepackage{color}

\usepackage[accsupp]{axessibility}  

\usepackage{subcaption}
\graphicspath{{images/}}

\usepackage{arydshln,booktabs}
\usepackage{multicol}

\begin{document}
\pagestyle{headings}
\mainmatter
\def\ECCVSubNumber{}  

\title{Learning visual explanations for DCNN-based image classifiers using an attention mechanism} 

\titlerunning{Learning-based CAM}
%
\author{Ioanna Gkartzonika \and
Nikolaos Gkalelis \and
Vasileios Mezaris}
\authorrunning{I. Gkartzonika et al.}
%
\institute{
CERTH-ITI, 6th Km Charilaou-Thermi Road, P.O. BOX 60361, Thessaloniki, Greece\\
\email{\{gkartzoni,gkalelis,bmezaris\}@iti.gr}}
\maketitle

\begin{abstract}
In this paper two new learning-based eXplainable AI (XAI) methods for deep convolutional
neural network (DCNN) image classifiers, called L-CAM-Fm and L-CAM-Img, are proposed.
Both methods use an attention mechanism that is inserted in the original (frozen) DCNN
and is trained to derive class activation maps (CAMs) from the last convolutional layer's feature maps.
During training, CAMs are applied to the feature maps (L-CAM-Fm) or the input image (L-CAM-Img)
forcing the attention mechanism to learn the image regions explaining the DCNN's outcome.
Experimental evaluation on ImageNet shows that the proposed methods
achieve competitive results while requiring a single forward pass at the inference stage.
Moreover, based on the derived explanations a comprehensive qualitative analysis
is performed providing valuable insight for understanding the reasons behind classification errors,
including possible dataset biases affecting the trained classifier.
\end{abstract}

\section{Introduction}
\label{INTRODUCTION}

During the last years, we are witnessing a breakthrough performance of DCNN image classifiers. 
However, the widespread commercial adoption of 
these methods is still hindered by the difficulty of users to attain some kind of explanations concerning the DCNN decisions.
This lack of DCNN transparency affects especially the adoption of this technology in safety-critical applications
as in the medical, security and self-driving vehicles industries,
where a wrong DCNN decision may have serious implications.
To this end, there is great demand for developing eXplainable AI (XAI) methods 
\cite{KimECCV2018,BaiPR2021,JungIcip2021,UeharaIcip2020,PrabhushankarIcip2020,JungECCV2020,PlummerECCV2020,MuddamsettyIcip2020,Hu_2022_WACV,HolzingerICMLW2020}.

A category of XAI approaches for DCNN image classifiers that is currently getting increasing attention
concerns methods that provide a visual explanation depicting the regions
of the input image that contribute the most to the decision of the classifier.
We should note that these approaches differ from methods used in weakly
supervised learning tasks such as weakly supervised object localization,
where the goal is to locate the region of the target object \cite{XiangweiICPR21,PengTaoICIP2021}.
This for instance can be seen in the explanation examples of the various figures
in our experimental evaluation section (Section \ref{SS:Evaluation}),
where often the focus region produced by the XAI approach does not
coincide with the region of the object instance corresponding to the class label of the image.

Gradient-based class activation map (CAM) \cite{CAM,Grad-CAM,Grad-CAM++,IntegratedGrad-CAM,Ablation-CAM}
and perturbation-based \cite{Score-CAM,RISE,SISE,ADA-SISE} approaches
have shown promising explanation performance.
Given an input image and its inferred class label, these methods generate a CAM,
which is rescaled to the image size providing the so-called saliency map (SM);
the SM indicates the image regions that the DCNN has focused on in order to infer this class.
However, these methods are either based on backpropagating gradients \cite{Grad-CAM,Grad-CAM++,IntegratedGrad-CAM},
producing suboptimal SMs due to the well-known gradient problems \cite{AdebayoNIPS18},
or require many forward passes at the inference stage \cite{Score-CAM,RISE,SISE,ADA-SISE},
thus introducing significant computational overhead.
Furthermore, the training dataset is not exploited in the exploration of the
internal mechanisms concerning the decision process of the classifier.
To this end, two new learning-based CAM methods are proposed,
called L-CAM-Fm and L-CAM-Img, which utilize an appropriate loss function
to train an attention mechanism \cite{BahdanauIclr15} for generating visual explanations.
Both methods can be used to generate explanations for arbitrary DCNN classifiers,
are gradient-free and during inference require only one forward pass
to derive a CAM and generate the respective SM of an input image.
Experimental evaluation using VGG-16 and ResNet-50 backbones on Imagenet shows
the efficacy of the proposed approaches in terms of both
explainability performance and computational efficiency.
Moreover, an extensive qualitative analysis using the generated SMs to explain
misclassification errors leads to interesting conclusions including, among others,
possible biases in the classifiers training dataset.
In summary, the contributions of this paper are:
\begin{itemize}
\item We present the first, to the best of our knowledge,
learning-based  CAM framework for explaining image classifiers;
this materializes into two XAI methods, L-CAM-Fm and L-CAM-Img.
\item An appropriate loss function, consisting of the cross-entropy loss
and an average and total variation CAM loss components,
is employed during training, forcing  the attention mechanism
to extract CAMs of low energy that constitute good explanations.
\end{itemize}

\begin{figure}[!htb]
\begin{center}
\begin{tabular}{cc}
{\small (a)} \hspace{0.5cm} &
\includegraphics[align=c,height=0.3\columnwidth]{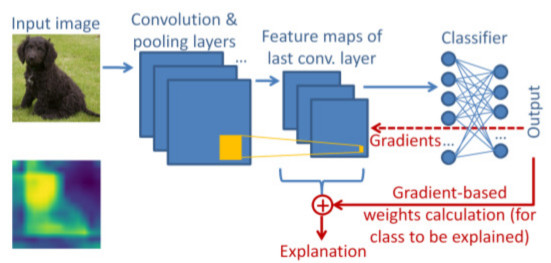} \\\\
{\small (b)} \hspace{0.5cm} &
\includegraphics[align=c,height=0.3\columnwidth]{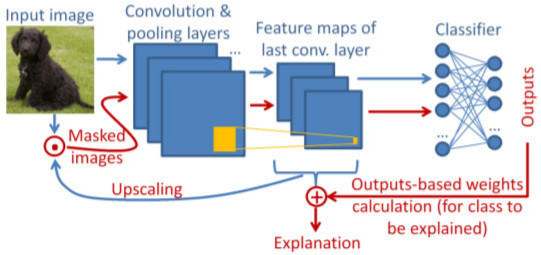} \\\\
{\small (c)} \hspace{0.5cm} &
\includegraphics[align=c,height=0.3\columnwidth]{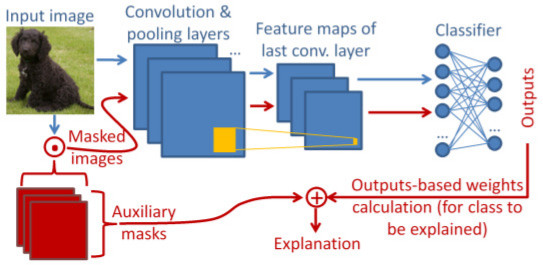}
\end{tabular}
\end{center}
\caption{Inference stage for literature XAI approaches:
(a) Gradient-based CAM methods, (b) Perturbation-based methods with feature maps, and,
(c) Perturbation-based methods with auxiliary masks.
Blue components and arrows denote the (frozen) DCNN classifier and the flow of the original input image through its layers.
Red arrows, indicating the flow of modified information such as masked images,
and components in red, are introduced by the respective approach to derive the model decision's explanation (SMs).
}
\label{fig:XAI_approaches}
\end{figure}


The paper is structured as follows:
The related work and proposed method are presented in
Sections \ref{RELATED_WORK} and \ref{PROPOSED_METHOD}.
Experimental results and conclusions are provided in Sections \ref{EXPERIMENTS} and \ref{CONCLUSION}.

\section{Related work}
\label{RELATED_WORK}

We discuss here visual XAI approaches that are mostly related to ours.
For a more comprehensive survey the reader is referred to \cite{BaiPR2021,WojciechSpringer2019,ArrietaIF20,HolzingerICMLW2020}.

\textit{Gradient-based CAM approaches} (Fig. \ref{fig:XAI_approaches}a) calculate a weight for each
feature map of the last convolutional layer using the gradients backpropagated from the output;
and, derive the CAM as the weighted sum of the feature maps \cite{Grad-CAM,Grad-CAM++,IntegratedGrad-CAM}.
Grad-CAM \cite{Grad-CAM} calculates the importance of each feature map
by the  gradients flowing  from the output layer into the last  convolutional  layer.
In \cite{Grad-CAM++}, Grad-CAM++ utilizes a weighted combination of the positive partial derivatives of the last convolutional layer.
Integrated Grad-CAM \cite{IntegratedGrad-CAM} introduces Integrated Gradient \cite{IntegratedGradient} to further improve the CAM's quality.

\textit{Perturbation-based approaches} 
are gradient-free
\cite{Score-CAM,RISE,SISE,ADA-SISE}.
Both Score-CAM \cite{Score-CAM} and SIDU \cite{MuddamsettyIcip2020} derive the weight of each feature map by forward passing perturbed copies of the input image.
Similarly, RISE \cite{RISE} generates randomly masked versions of the input image to compute the aggregation weights.
In \cite{SISE}, SISE selects feature maps at various depths, generates the so-called attribution masks and combines them using their classification scores.
In \cite{ADA-SISE}, a fraction of the feature maps are adaptively selected by ADA-SISE, reducing the computational complexity of SISE.
The general architecture of SIDU, Score-CAM, SISE and ADA-SISE is shown in Fig. \ref{fig:XAI_approaches}b,
while the respective architecture for RISE is depicted in Fig. \ref{fig:XAI_approaches}c.
The form of explanation (SM) produced by all methods is shown right below the input image in Fig. \ref{fig:XAI_approaches}a.

\section{Proposed method}
\label{PROPOSED_METHOD}

\subsection{Problem formulation}\label{PROBLEM_FORMULATION}

Let $f$ be a DCNN model trained to categorize images to one of $R$ different classes.
Suppose an input image $\mathbf{X} \in \mathbb{R}^{W \times H \times C}$
that passes through $f$ producing a model-truth label $y \in \{1,\dots,R\}$,
i.e. the top-1 class label inferred by $f$, and $K$ feature maps extracted from $f$'s last convolutional layer,
\begin{equation}
\mathbf{A} \in \mathbb{R}^{P \times Q \times K},
\end{equation}
where, $W$, $H$, $C$ and $P$, $Q$, $K$, are the width,
height and number of channels of $\mathbf{X}$ and $\mathbf{A}$, respectively,
and $\mathbf{A}_{:,:,k}$ is the $k$th feature map.
Given the above formulation, the goal of CAM-based methods is to derive an activation map from the $K$ feature maps,
the so-called CAM, and based on  it generate the respective SM, visualizing the salient image regions that explain $f$'s decision.

\begin{figure}[!htb]
\begin{center}
\begin{tabular}{cc}
{\small (a)} \hspace{0.5cm} &
\includegraphics[align=c,height=0.3\columnwidth]{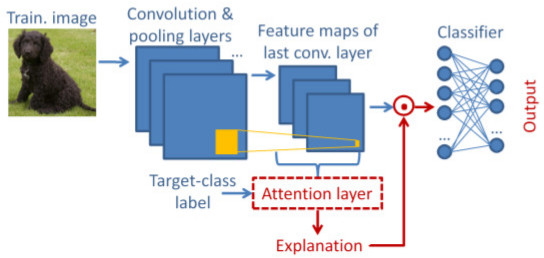} \\\\
{\small (b)} \hspace{0.5cm} &
\includegraphics[align=c,height=0.3\columnwidth]{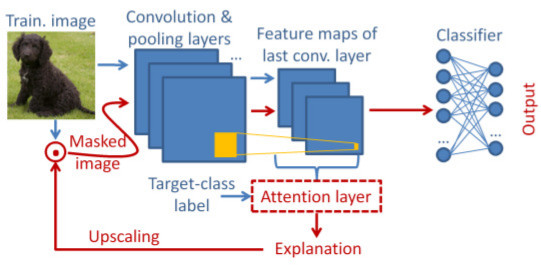} \\\\
{\small (c)} \hspace{0.5cm} &
\includegraphics[align=c,height=0.3\columnwidth]{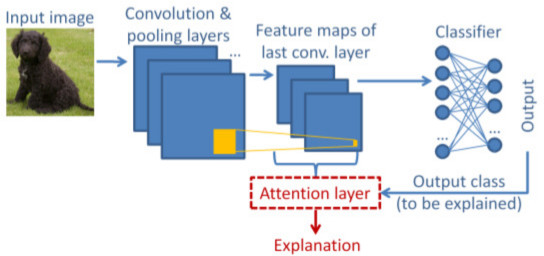}
\end{tabular}
\end{center}
\caption{Network architectures of the proposed approaches: (a) L-CAM-Fm training, (b) L-CAM-Img training, (c) L-CAM-Fm/-Img inference.
}
\label{fig:proposedApproach}
\end{figure}

\subsection{Training the attention mechanism}
\label{SS:Training}

Consider a training set of $R$ classes (the same classes that were used to train $f$),
where each image $\mathbf{X}$ in the dataset is associated with a model-truth label $y$.
This dataset is used to train an attention mechanism $g()$
\begin{equation}
    \mathbf{L}^{(y)} = g(y, \mathbf{A}), \label{E:CAM}
\end{equation}
where $\mathbf{L}^{(y)} \in \mathbb{R}^{P \times Q}$ is the CAM produced for a specified $\mathbf{X}$ and $y$.
Specifically, the attention mechanism is implemented as follows
\begin{equation}
    g(y, \mathbf{A}) = \sum_{k=1}^{K} w_k^{(y)} \mathbf{A}_{:,:,k} + b^{(y)} \mathbf{J}, \label{E:FeedForwardNet}
\end{equation}
where, the weight matrix $\mathbf{W} = [\mathbf{w}^{(1)}, \dots, \mathbf{w}^{(R)}]^T$ and
bias vector $\mathbf{b} = [b^{(1)}, \dots, b^{(R)}]^T$ are the parameters of the attention mechanism,
$\mathbf{w}^{(r)} = [w_1^{(r)}, \dots, w_K^{(r)}]^T$ is the $r$th row of $\mathbf{W}$,
$w_k^{(r)} \in \mathbb{R}$ is the $k$th element of $\mathbf{w}^{(r)}$,
and $\mathbf{J} \in \mathbb{R}^{P \times Q}$ is an all-ones matrix.
That is, the model-truth label $y$ at the input of $g()$ is used to select the class-specific weight vector
and bias term from the $y$th row of $\mathbf{W}$ and $\mathbf{b}$, respectively.

To learn the parameters of the attention mechanism, two different approaches,
called L-CAM-Fm and L-CAM-Img, are proposed,
with the corresponding network architectures shown in Figs. \ref{fig:proposedApproach}a and \ref{fig:proposedApproach}b.
In both architectures, the attention mechanism is placed at the output of the last convolutional layer of the DCNN
and the elements of the derived CAM are normalized to [0,1] using the element-wise sigmoid function $\sigma()$.
In L-CAM-Fm, the CAM produced by the attention mechanism is used as
a self-attention mask to re-weight the elements of the feature maps, i.e., 
\begin{equation}
    \mathbf{A}_{:,:,k} \leftarrow \mathbf{A}_{:,:,k} \odot \sigma(\mathbf{L}^{(y)}), k=1, \dots, K, \label{E:LCAMFm}
\end{equation}
where $\odot$ denotes element-wise multiplication.
Contrarily, in L-CAM-Img the derived CAM is upscaled and applied to each channel of the input image,
\begin{equation}
    \mathbf{X}_{:,:,c} \leftarrow \mathbf{X}_{:,:,c} \odot \theta(\sigma(\mathbf{L}^{(y)})), c=1,\dots,C, \label{E:LCAMImg}
\end{equation}
where, $\theta(): \mathbb{R}^{P \times Q} \rightarrow \mathbb{R}^{W \times H}$
is the upscaling operator (e.g. bilinear interpolation) and $\mathbf{X}_{:,:,c}$ is the $c$th channel of $\mathbf{X}$.

The overall architecture is trained end-to-end using an iterative gradient descent algorithm,
where the attention mechanism's weights are updated at every iteration, while the weights of  $f$ remain fixed to their original values.
The following loss function is used during training for both L-CAM-Fm and L-CAM-Img,
\begin{equation}
\lambda_{1} TV(\sigma(\mathbf{L}^{(y)})) + \lambda_{2} AV(\sigma(\mathbf{L}^{(y)})) + \lambda_{3} CE(y,u), \label{E:Loss}
\end{equation}
where, $CE(,)$ is the cross-entropy loss, $u$ is the confidence score for class $y$ derived
using the L-CAM-Fm or -Img network, $\lambda_{1}$, $\lambda_{2}$, $\lambda_{3}$,
are regularization parameters, and $AV()$, $TV()$ are the average and total variation operator, respectively.
For the two latter operators we use the definitions presented in \cite{DabkowskiNIPS17},
\begin{eqnarray}
 AV(\mathbf{S}) &=& \frac{1}{P Q} \sum_{p,q} (s_{p,q})^{\lambda_4}, \label{E:AvLoss} \\
  TV(\mathbf{S}) &=& \sum_{p,q} [(s_{p,q} - s_{p,q+1})^2 + (s_{p,q} - s_{p+1,q})^2], \label{E:TvLoss}
\end{eqnarray}
where, $s_{p,q}$ is the element at the $p$th row and $q$th column of
an arbitrary tensor $\mathbf{S} \in \mathbb{R}^{P \times Q}$ and $\lambda_4$ is a fourth regularization parameter.
We should note that although the weights of $f$ are kept frozen, the gradients backpropagate
through it and train effectively the attention mechanism parameters, as explained for instance in \cite{FrozenNIPS2021}.
Thus, the attention mechanism is forced to learn a transformation of the feature maps so that the CAM retains the regions of the input image that best explain $f$'s decision.

\subsection{Inference of model decision's explanation}

During the inference stage, the procedure to derive the CAM of a test image is the same for both L-CAM-Fm and L-CAM-Img (see Fig. \ref{fig:proposedApproach}c).
That is, the test image is forward-passed through the DCNN to produce
the corresponding feature maps and the model-truth label,
which are then forwarded to the trained attention mechanism for computing the CAM (Eqs. (\ref{E:CAM}), (\ref{E:FeedForwardNet})).
Similarly to \cite{Grad-CAM,Grad-CAM++}, the explanation (SM) $\mathbf{V} \in \mathbb{R}^{W \times H}$
is then derived by using a min-max normalization operator $\varsigma()$ and upscaling $\theta()$ to the input image size,
\begin{equation}
    \mathbf{V}^{(y)} = \theta(\varsigma(\mathbf{L}^{(y)})). \label{E:LcamInfer}
\end{equation}

\section{Experiments}
\label{EXPERIMENTS}

\subsection{Dataset}

ImageNet \cite{ImageNet}, which is among the most popular datasets in the visual XAI domain, was selected for the experiments.
It contains $R=1000$ classes, ~1.3 million images for training and 50K images for testing.
Due to the prohibitively high computational cost of perturbation-based approaches that are used
for experimental comparison, only 2000 randomly-selected testing images are used for evaluation,
following an evaluation protocol similar to \cite{Score-CAM}.

\subsection{Experimental setup}

The proposed L-CAM-Fm and L-CAM-Img are compared against the top-performing approaches
in the literature for which publicly-available code is provided, i.e.,
Grad-CAM \cite{Grad-CAM}, Grad-CAM++ \cite{Grad-CAM++}, Score-CAM \cite{Score-CAM},
and RISE \cite{SISE} (using the implementations of \cite{ScoreCamPytorch} for the first three and of \cite{RiseCode} for the fourth).
Two sets of experiments are conducted with respect to the employed DCNN classifier, i.e. one using VGG-16 \cite{VGG16} and another using ResNet-50 \cite{RESNET}.
In both cases, pretrained models from the PyTorch model zoo \cite{TorchvisionModels} are used.

The proposed approaches are trained using the loss of Eq. (\ref{E:Loss})
with stochastic gradient descent, batch size 64 and learning rate $10^{-4}$.
The learning rate decay factor per epoch and total number of epochs are 0.75, 7 for the VGG-16 experiment and 0.95, 25 for the ResNet-50 one.
The loss regularization parameters (Eqs. (\ref{E:Loss}), (\ref{E:AvLoss})) are chosen empirically using the training set
in order to minimize the total loss (Eq. (\ref{E:Loss})) and at the same time
bring the different loss components at the same order of magnitude
(thus ensuring that all of them contribute similarly to the loss function):
$\lambda_1 = 0.01$, $\lambda_2 = 2$, $\lambda_3 = 1.5$, $\lambda_4 = 0.3$.
We should note that in all experiments the proposed methods exhibit a quite
stable performance with respect to the above optimization parameters.
During training, each image is rescaled and normalized as done during training
of the original DCNN classifier, i.e. its shorter side is scaled to 256 pixels,
then random-cropped to $W \times H \times C$, where,
$W=H=224$ and $C=3$ (the three RGB channels) and normalized to zero mean and unit variance.
The same preprocessing is used during testing, except that center-cropping is applied.
The size $P \times Q \times K$ of the feature maps tensor at the last convolutional layer
of the DCNN is $P=Q=14$, $K=512$ and $P=Q=7$, $K=2048$ for VGG-16 and ResNet-50, respectively.
For the compared CAM approaches the SM of an input image is derived as follows \cite{Grad-CAM,Grad-CAM++}:
the derived CAM is normalized to [0,1] using the min-max operator and transformed
to the size of the input image using bilinear interpolation (Eq. \ref{E:LcamInfer}).
In contrary, for Score-CAM and RISE, as proposed in their corresponding papers \cite{Score-CAM,SISE},
the opposite procedure is followed to derive the SM, i.e.,
bilinear interpolation to the input image's size and then min-max normalization.

\begin{table*}
\centering
\begin{tabular}{l|cc:cc:cc:c}
  & \scriptsize AD(100\%)$\downarrow$  & \scriptsize IC(100\%)$\uparrow$ & \scriptsize AD(50\%)$\downarrow$  & \scriptsize IC(50\%)$\uparrow$ & \scriptsize AD(15\%)$\downarrow$ & \scriptsize IC(15\%)$\uparrow$ & \scriptsize \#FW$\downarrow$  \\ 
\hline
 Grad-CAM \cite{Grad-CAM}     & 32.12 & 22.1 & 58.65 & 9.5  & 84.15 &  2.2 & \textbf{1} \\
Grad-CAM++ \cite{Grad-CAM++} & 30.75 &  22.05 & 54.11 & 11.15 & 82.72 &  3.15  & \textbf{1} \\
Score-CAM \cite{Score-CAM}    & 27.75 & 22.8 &  45.6  &  14.1  & \underline{75.7} & \underline{4.3}  & 512 \\
RISE \cite{RISE}              & \textbf{8.74} &   \textbf{51.3} & \underline{42.42} & \underline{17.55} &  78.7 &  \textbf{4.45} &  4000  \\
L-CAM-Fm*    &  20.63 & 31.05 &  51.34 &  13.45 &  82.4 & 3.05 & \textbf{1}\\
L-CAM-Fm      & 16.47 &  35.4 & 47 &  14.45 & 79.39 &  3.65 &  \textbf{1} \\
L-CAM-Img*   & 18.01 &  37.2 &  50.88 &  12.05 & 82.1 & 3  &  \textbf{1}  \\
L-CAM-Img     & 12.96  &  \underline{41.25}  & 45.56   &  14.9 & 78.14  & 4.2 &  \textbf{1} \\
L-CAM-Img$^\dagger$ & \underline{12.15} &  40.95 & \textbf{37.37}  &  \textbf{20.25} & \textbf{74.23} & \textbf{4.45} & \textbf{1} \\
\hline
Grad-CAM \cite{Grad-CAM}      & 13.61  &  38.1 & 29.28 & 23.05 &  \textbf{78.61} & 3.4 & \textbf{1} \\
Grad-CAM++ \cite{Grad-CAM++} & 13.63 &  37.95 &30.37   &  23.45 & 79.58 & 3.4  & \textbf{1} \\
Score-CAM \cite{Score-CAM}    & \textbf{11.01} & 39.55 & \textbf{26.8} &  \textbf{24.75}  &  78.72  & 3.6 & 2048\\
RISE \cite{RISE}              & 11.12 &  \textbf{46.15} & 36.31 & 21.55 &  82.05  & 3.2 & 8000 \\
L-CAM-Fm*    & 14.44  &  35.45 & 32.18 &  20.5 &  80.66  & 2.9 &  \textbf{1} \\
L-CAM-Fm      &  12.16 &  40.2  & 29.44 & 23.4  &  \underline{78.64} &  \textbf{4.1} &  \textbf{1}  \\
L-CAM-Img*   &15.93  &  32.8  &  39.9  & 14.85 &  84.67 &  2.25 &  \textbf{1}  \\
L-CAM-Img    &  \underline{11.09} & \underline{43.75} &  \underline{29.12} &  \underline{24.1} & 79.41 &  \underline{3.9} &  \textbf{1}
\end{tabular}
\caption{\label{tab:Results} Evaluation results for a VGG-16 (upper half) and ResNet-50 (lower half)
backbone classifier using 2000 randomly-selected testing images of ImageNet.
The best and 2nd-best performance for a given evaluation measure are shown in bold and underline, respectively.}
\end{table*}

\begin{figure}[!htb]
\begin{center}
\begin{tabular}{ccccccc}
\rotatebox[]{90}{\scriptsize Image} &
\includegraphics[align=c,width=0.15\columnwidth]{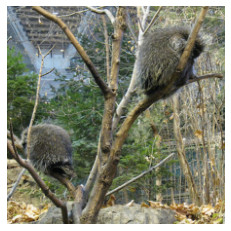} &
\includegraphics[align=c,width=0.15\columnwidth]{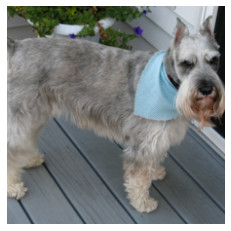} &
\includegraphics[align=c,width=0.15\columnwidth]{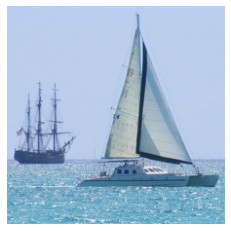} &
\includegraphics[align=c,width=0.15\columnwidth]{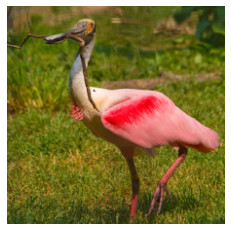} &
\includegraphics[align=c,width=0.15\columnwidth]{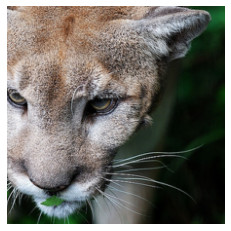} &
\includegraphics[align=c,width=0.15\columnwidth]{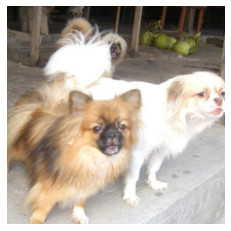} \\
\rotatebox[]{90}{\scriptsize Grad-CAM} &
\includegraphics[align=c,width=0.15\columnwidth]{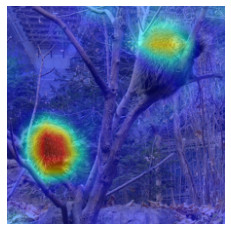} &
\includegraphics[align=c,width=0.15\columnwidth]{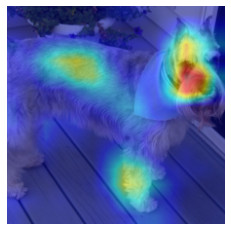} &
\includegraphics[align=c,width=0.15\columnwidth]{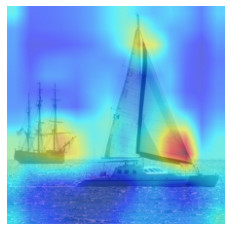} &
\includegraphics[align=c,width=0.15\columnwidth]{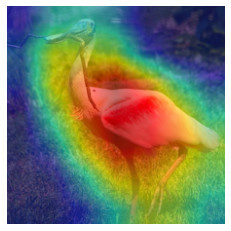} &
\includegraphics[align=c,width=0.15\columnwidth]{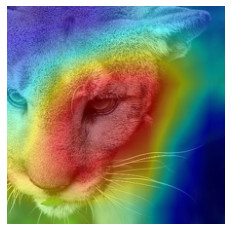} &
\includegraphics[align=c,width=0.15\columnwidth]{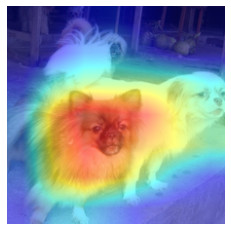} \\
\rotatebox[]{90}{\scriptsize Grad-CAM++} &
\includegraphics[align=c,width=0.15\columnwidth]{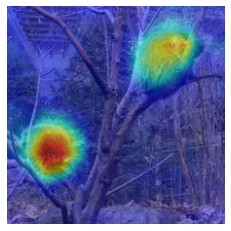} &
\includegraphics[align=c,width=0.15\columnwidth]{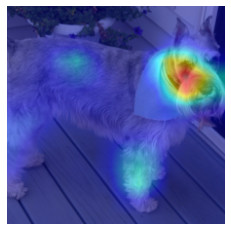} &
\includegraphics[align=c,width=0.15\columnwidth]{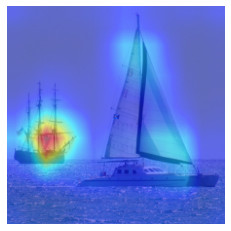} &
\includegraphics[align=c,width=0.15\columnwidth]{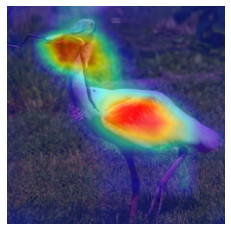} &
\includegraphics[align=c,width=0.15\columnwidth]{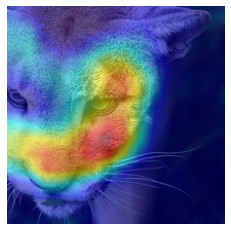} &
\includegraphics[align=c,width=0.15\columnwidth]{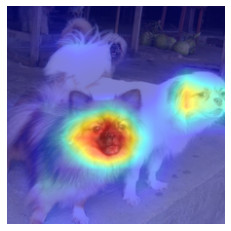}  \\
\rotatebox[]{90}{\scriptsize Score-CAM} &
\includegraphics[align=c,width=0.15\columnwidth]{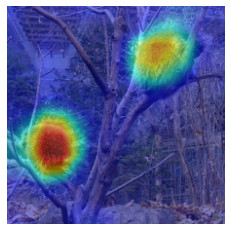} &
\includegraphics[align=c,width=0.15\columnwidth]{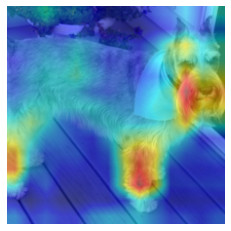} &
\includegraphics[align=c,width=0.15\columnwidth]{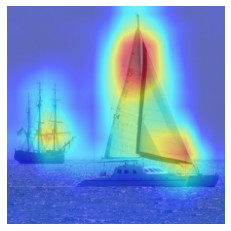} &
\includegraphics[align=c,width=0.15\columnwidth]{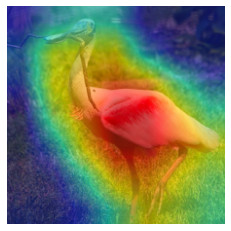} &
\includegraphics[align=c,width=0.15\columnwidth]{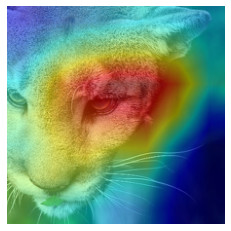} &
\includegraphics[align=c,width=0.15\columnwidth]{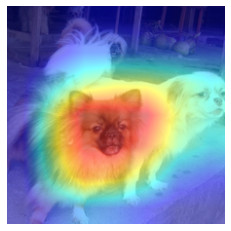} \\
\rotatebox[]{90}{\scriptsize RISE} &
\includegraphics[align=c,width=0.15\columnwidth]{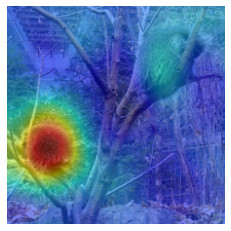} &
\includegraphics[align=c,width=0.15\columnwidth]{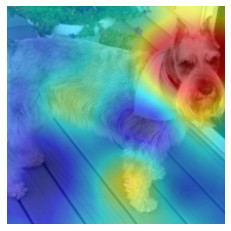} &
\includegraphics[align=c,width=0.15\columnwidth]{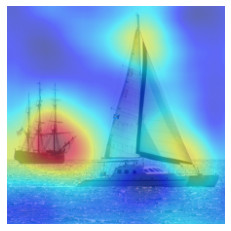} &
\includegraphics[align=c,width=0.15\columnwidth]{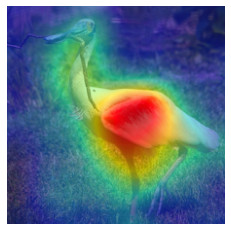} &
\includegraphics[align=c,width=0.15\columnwidth]{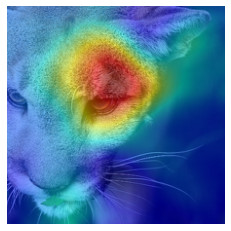} &
\includegraphics[align=c,width=0.15\columnwidth]{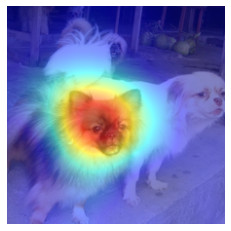} \\
\rotatebox[]{90}{\scriptsize L-CAM-Fm}  &
\includegraphics[align=c,width=0.15\columnwidth]{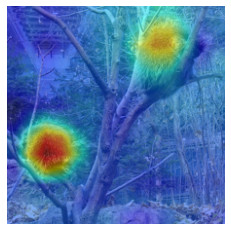} &
\includegraphics[align=c,width=0.15\columnwidth]{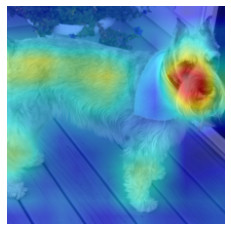} &
\includegraphics[align=c,width=0.15\columnwidth]{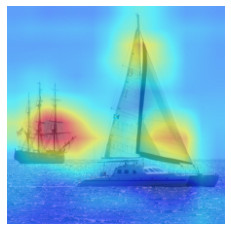} &
\includegraphics[align=c,width=0.15\columnwidth]{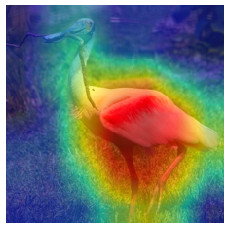} &
\includegraphics[align=c,width=0.15\columnwidth]{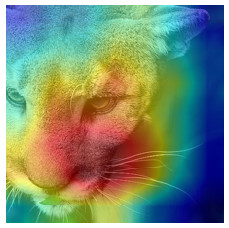} &
\includegraphics[align=c,width=0.15\columnwidth]{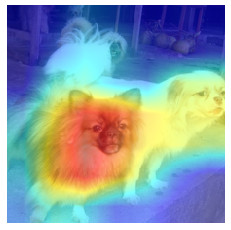} \\
\rotatebox[]{90}{{\scriptsize L-CAM-Img}} &
\includegraphics[align=c,width=0.15\columnwidth]{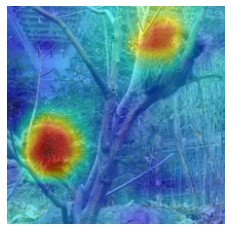} &
\includegraphics[align=c,width=0.15\columnwidth]{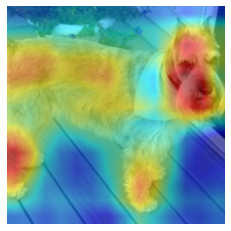} &
\includegraphics[align=c,width=0.15\columnwidth]{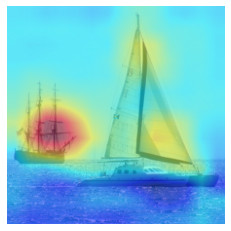} &
\includegraphics[align=c,width=0.15\columnwidth]{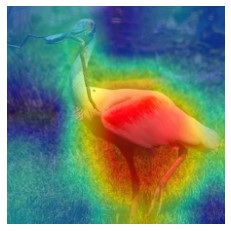} &
\includegraphics[align=c,width=0.15\columnwidth]{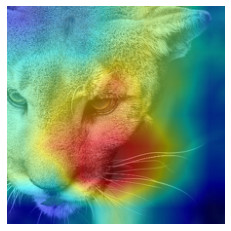} &
\includegraphics[align=c,width=0.15\columnwidth]{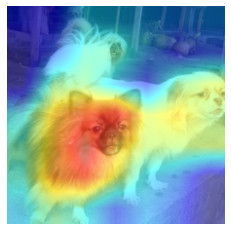} 
\end{tabular}
\end{center}
\caption{Visualization of SMs with $\nu = 100\%$ from various XAI methods superimposed
on the original input image to produce class-specific visual explanations
for the VGG-16 (columns 1 to 3) and ResNet-50 (columns 4 to 6) backbones.
}
\label{fig:explanationTwoImg}
\end{figure}

\begin{figure}[!htb]
\begin{center}
\begin{tabular}{ccc}
\includegraphics[width=0.23\columnwidth]{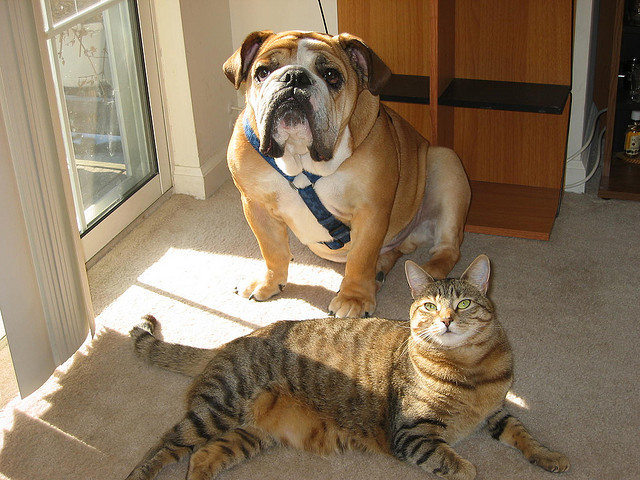} &
\includegraphics[width=0.23\columnwidth]{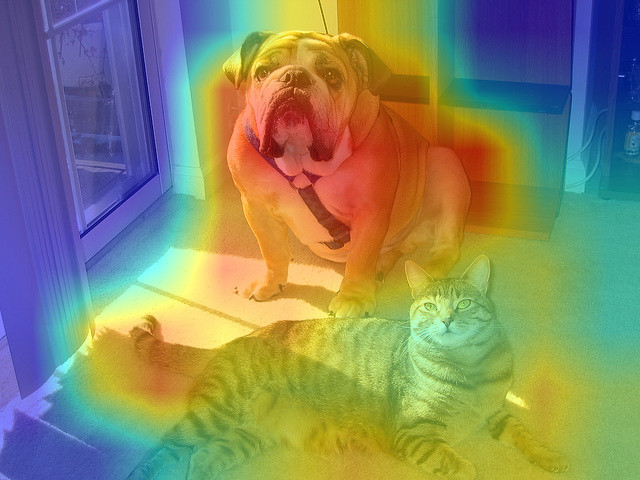} & 
\includegraphics[width=0.23\columnwidth]{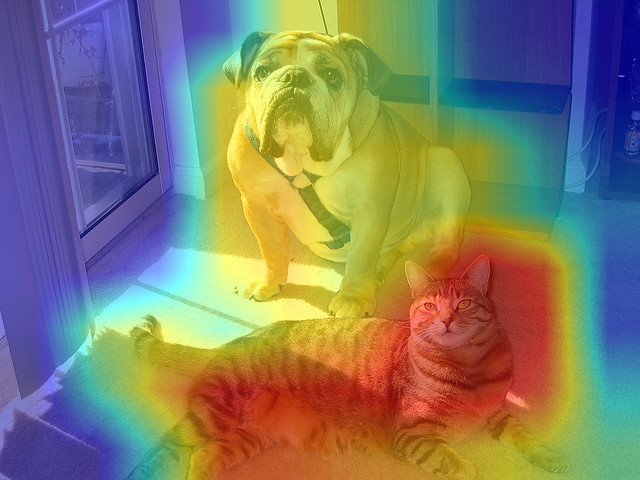} \\
{\scriptsize Image}& 
{\scriptsize Pug} &
{\scriptsize Tiger cat}  \\\\
\includegraphics[width=0.23\columnwidth]{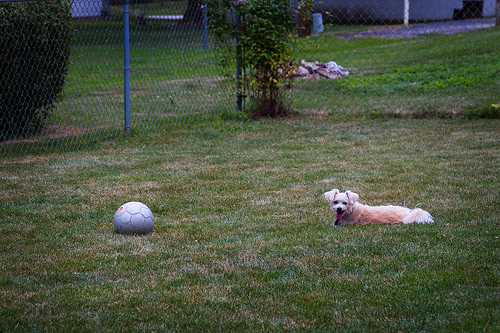}&
\includegraphics[width=0.23\columnwidth]{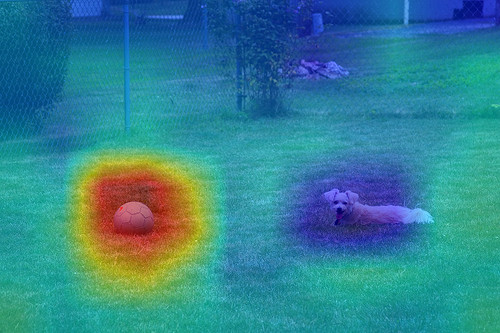} &
\includegraphics[width=0.23\columnwidth]{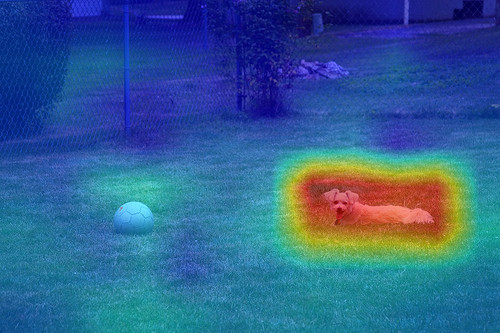} \\
{\scriptsize Image} &  
{\scriptsize soccer ball} &  
{\scriptsize Maltese}
\end{tabular}
\end{center}
\caption{Two examples of using class-specific SMs (superimposed on the input image) produced by L-CAM-Img$^\dagger$
on VGG-16 with $\nu = 100\%$ for classes ``pug'' and ``tiger cat'' (top row)
and classes ``soccer ball'' and ``Maltese'' (bottom row).}
\label{fig:explanationClassSpecific}
\end{figure}

\subsection{Evaluation measures}

Two widely used evaluation measures, Increase in Confidence (IC) and Average Drop (AD) \cite{Grad-CAM++}, are used in the experimental evaluation,

\begin{eqnarray}
\mbox{AD} &=& \sum_{i=1}^{\Upsilon} \frac{max(0, f(\mathbf{X}_i)- f(\mathbf{X}_i \odot \phi_{\nu}(\mathbf{V}_i)))}{\Upsilon f(\mathbf{X}_i)} 100, \label{E:IC}\\
\mbox{IC} &=& \sum_{i=1}^{\Upsilon} \frac{\delta(f(\mathbf{X}_i \odot \phi_{\nu}(\mathbf{V}_i)) > f(\mathbf{X}_i))}{\Upsilon} 100, \label{E:IC}
\end{eqnarray}
where $f()$ is the original DCNN classifier, $\phi_{\nu}()$ is a threshold function to select
the $\nu$ percent higher-valued pixels of the SM \cite{Ablation-CAM,Score-CAM},
$\delta()$ returns 1 when the input condition is satisfied and zero otherwise,
$\Upsilon$ is the number of test images, $\mathbf{X}_i$ is the $i$th test image
and $\mathbf{V}_i$ is the respective SM produced by the XAI method under evaluation.

\subsection{Results}
\label{SS:Evaluation}

\begin{figure}[!htb]
\begin{center}
\begin{tabular}{cccccc}
\rotatebox[]{90}{\scriptsize Image} &
\includegraphics[align=c,height=0.16\columnwidth]{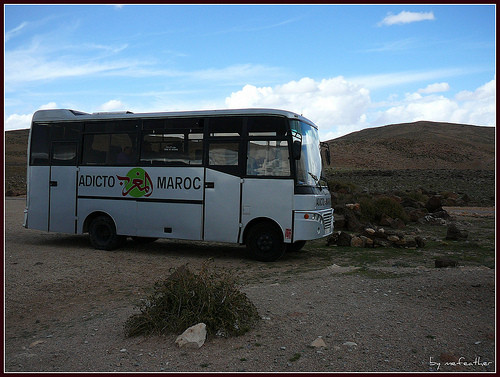} &
\includegraphics[align=c,height=0.16\columnwidth]{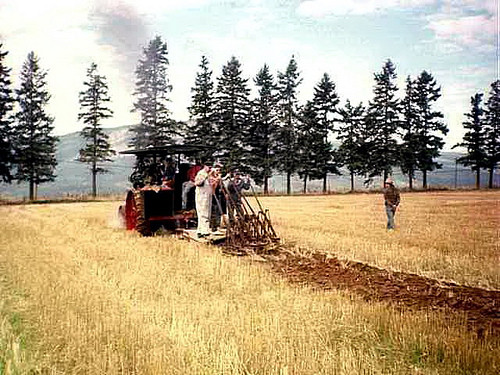} &
\includegraphics[align=c,height=0.16\columnwidth]{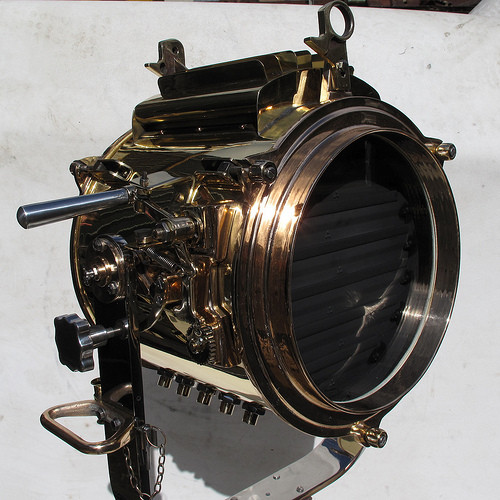} &
\includegraphics[align=c,height=0.16\columnwidth]{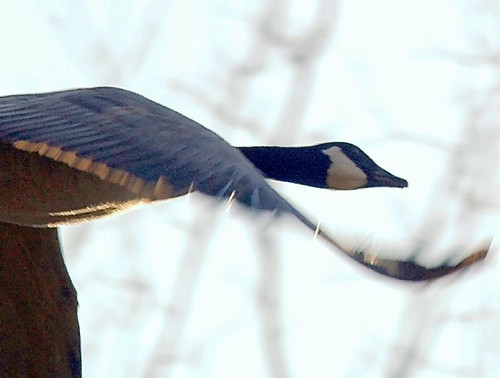} &
\includegraphics[align=c,height=0.16\columnwidth]{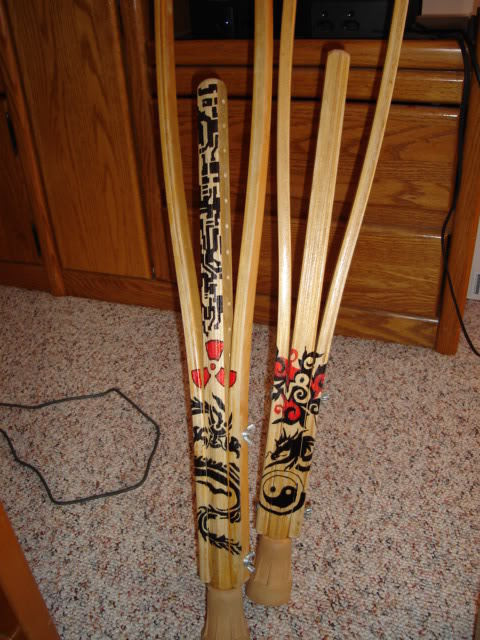} \vspace{0.08cm} \\
\rotatebox[]{90}{\scriptsize Ground truth} &
\includegraphics[align=c,height=0.16\columnwidth]{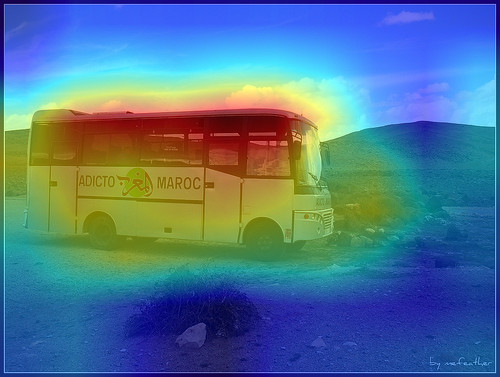} &
\includegraphics[align=c,height=0.16\columnwidth]{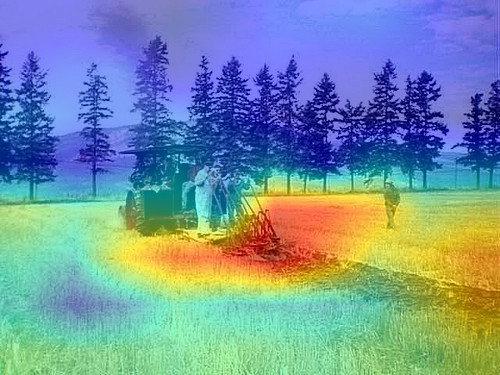} &
\includegraphics[align=c,height=0.16\columnwidth]{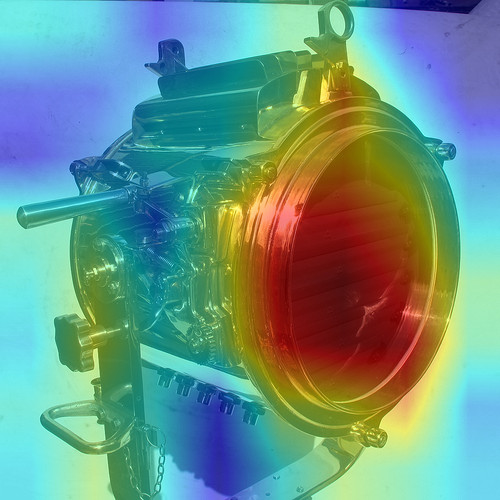} &
\includegraphics[align=c,height=0.16\columnwidth]{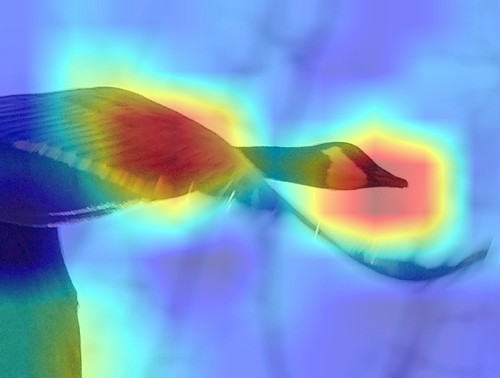} &
\includegraphics[align=c,height=0.16\columnwidth]{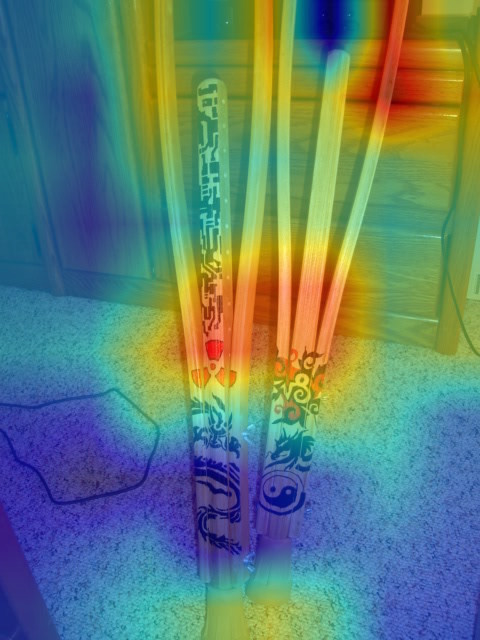} \vspace{0.08cm} \\
 &
{\scriptsize coach} &
{\scriptsize plow} &
{\scriptsize spotlight} &
{\scriptsize goose} &
{\scriptsize crutch} \\\\
\rotatebox[]{90}{{\scriptsize Predicted}} &
\includegraphics[align=c,height=0.16\columnwidth]{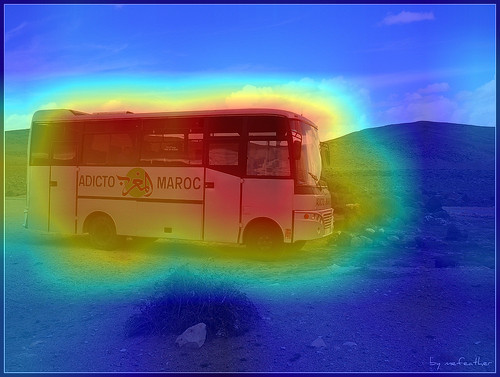} &
\includegraphics[align=c,height=0.16\columnwidth]{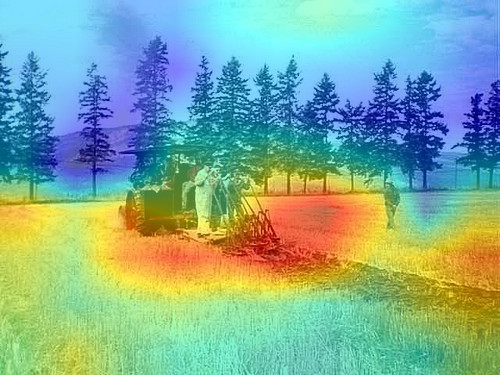} &
\includegraphics[align=c,height=0.16\columnwidth]{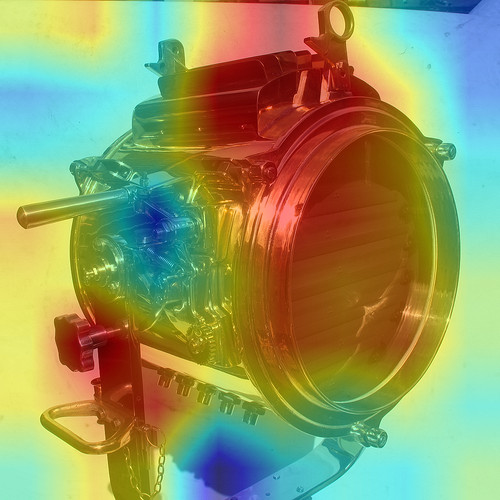} &
\includegraphics[align=c,height=0.16\columnwidth]{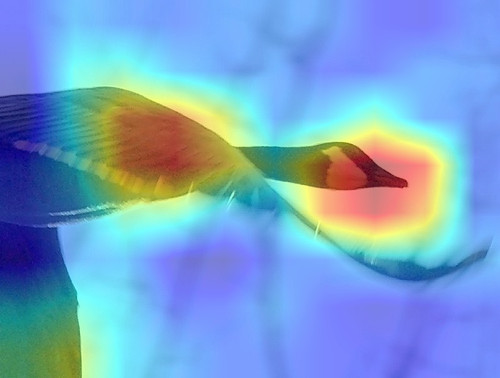} &
\includegraphics[align=c,height=0.16\columnwidth]{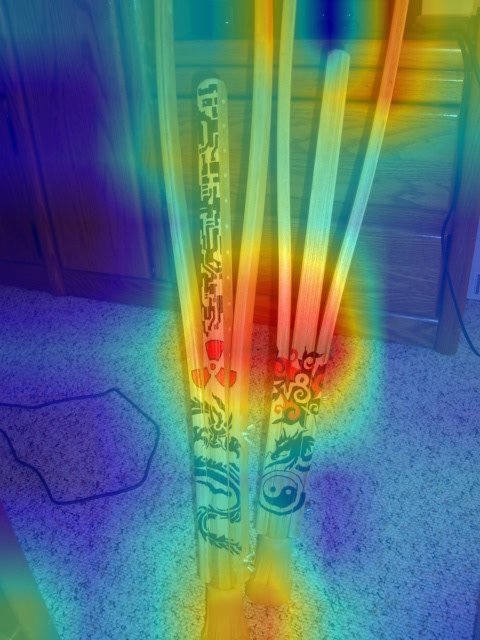} \vspace{0.08cm} \\
 &
{\scriptsize minibus} &
{\scriptsize thresher}  &
{\scriptsize projector} &
{\scriptsize drake} &
{\scriptsize drumstick}
\end{tabular}
\end{center}
\caption{Illustration of images and class-specific SMs (superimposed on the input image)
whose ground truth and predicted labels are highly correlated.
}
\label{fig:similarClasses}
\end{figure}

\begin{figure}[!htb]
\begin{center}
\begin{tabular}{cccccc}
\rotatebox[]{90}{\scriptsize Image} &
\includegraphics[align=c,height=0.16\columnwidth]{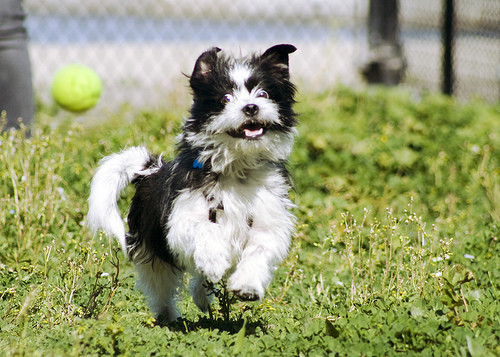} &
\includegraphics[align=c,height=0.16\columnwidth]{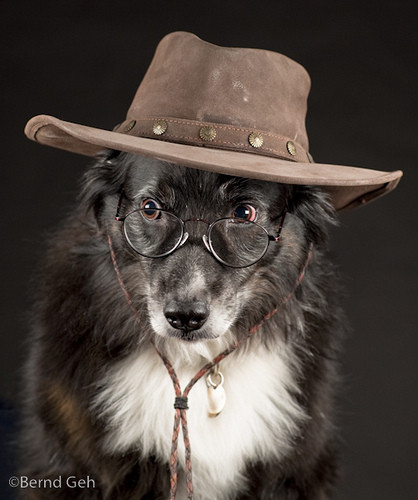} &
\includegraphics[align=c,height=0.16\columnwidth]{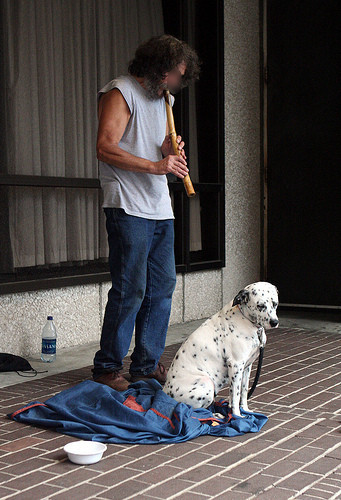} &
\includegraphics[align=c,height=0.16\columnwidth]{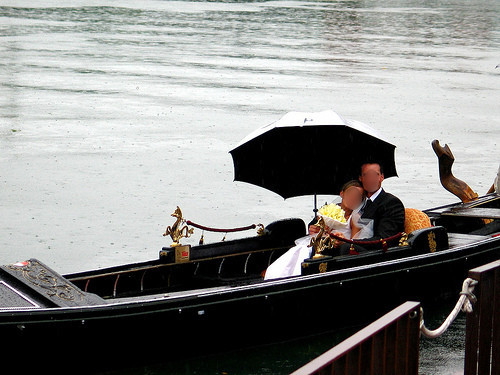} &
\includegraphics[align=c,height=0.16\columnwidth]{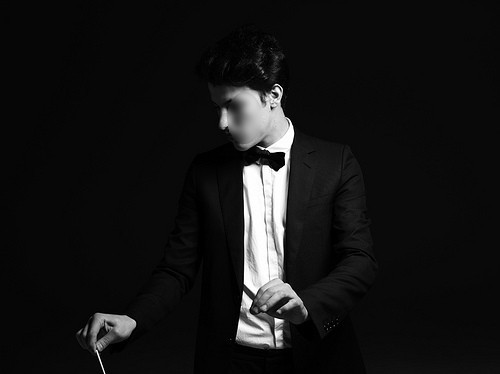} \vspace{0.08cm} \\
\rotatebox[]{90}{\scriptsize Ground truth} &
\includegraphics[align=c,height=0.16\columnwidth]{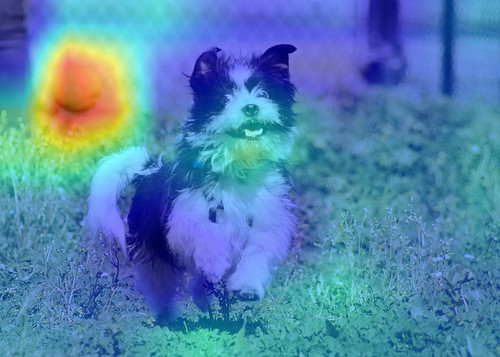} &
\includegraphics[align=c,height=0.16\columnwidth]{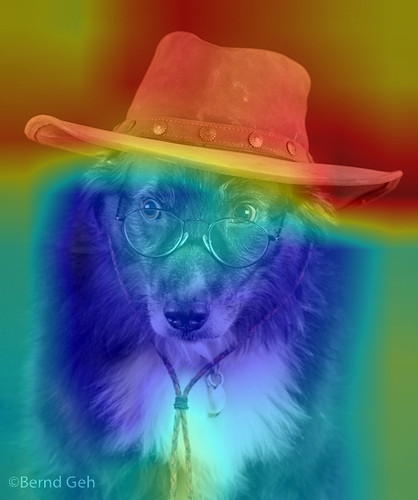} &
\includegraphics[align=c,height=0.16\columnwidth]{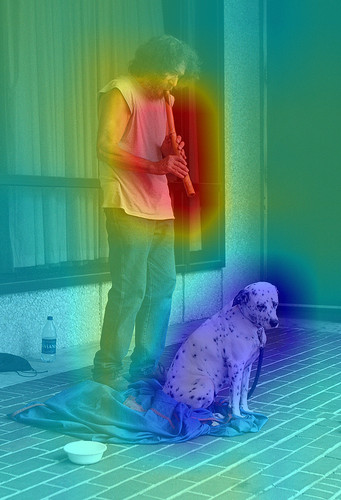} &
\includegraphics[align=c,height=0.16\columnwidth]{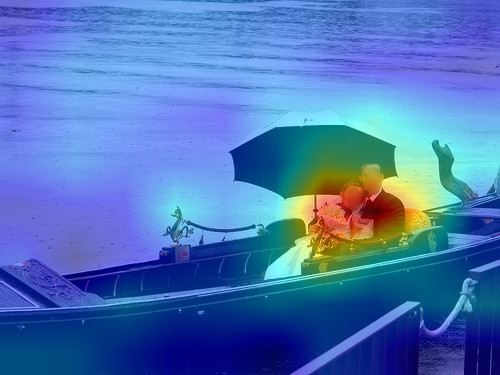} &
\includegraphics[align=c,height=0.16\columnwidth]{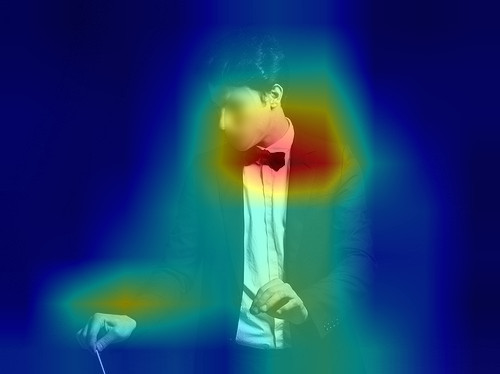} \vspace{0.08cm} \\
 &
{\scriptsize tennis ball} &
{\scriptsize cowboy hat} &
{\scriptsize flute} &
{\scriptsize groom} &
{\scriptsize bow tie} \\\\
\rotatebox[]{90}{{\scriptsize Predicted}} &
\includegraphics[align=c,height=0.16\columnwidth]{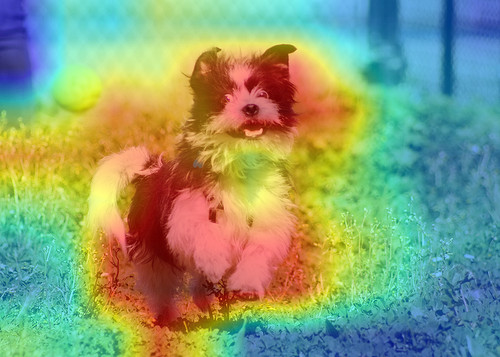} &
\includegraphics[align=c,height=0.16\columnwidth]{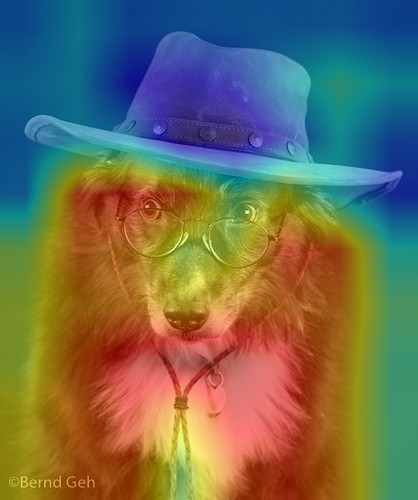} &
\includegraphics[align=c,height=0.16\columnwidth]{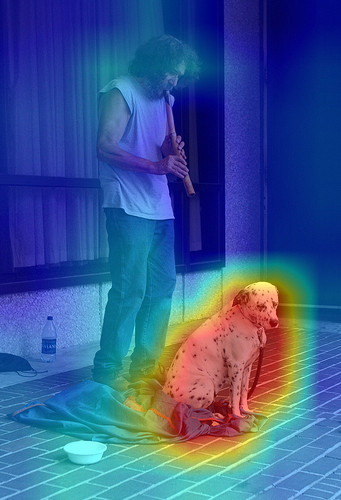} &
\includegraphics[align=c,height=0.16\columnwidth]{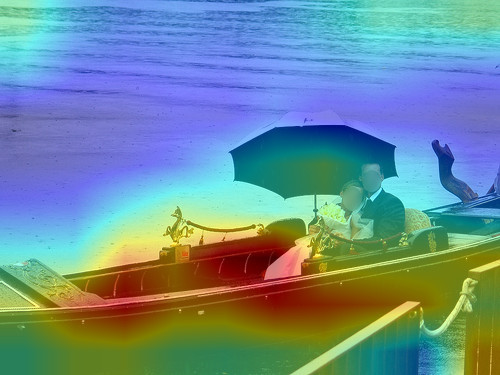} &
\includegraphics[align=c,height=0.16\columnwidth]{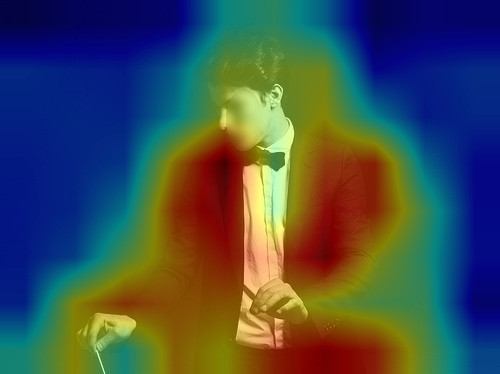} \vspace{0.08cm} \\
 &
{\scriptsize Tibetan terrier} &
{\scriptsize Border collie}  &
{\scriptsize Dalmatian} &
{\scriptsize gondola} &
{\scriptsize suit}
\end{tabular}
\end{center}
\caption{Illustration of images and class-specific SMs (superimposed on the input image),
which (although single-labelled) contain instances of two different Imagenet classes.
}
\label{fig:multilabelSamples}
\end{figure}

\begin{figure}[!htb]
\begin{center}
\begin{tabular}{cccccc}
\rotatebox[]{90}{\scriptsize Image} &
\includegraphics[align=c,height=0.18\columnwidth]{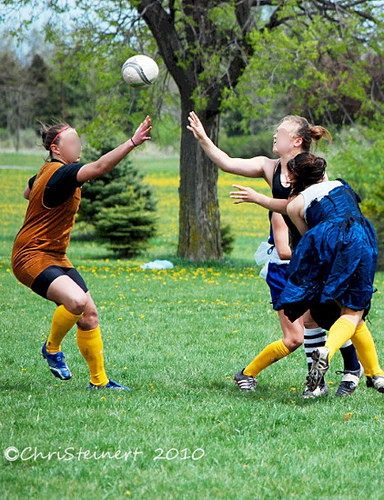} &
\includegraphics[align=c,height=0.18\columnwidth]{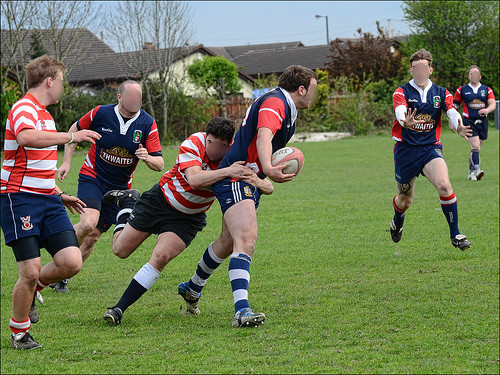} &
\includegraphics[align=c,height=0.18\columnwidth]{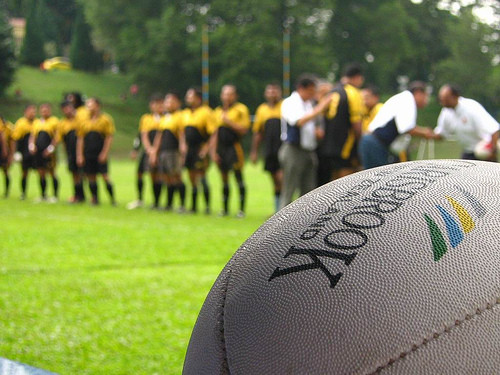} &
\includegraphics[align=c,height=0.18\columnwidth]{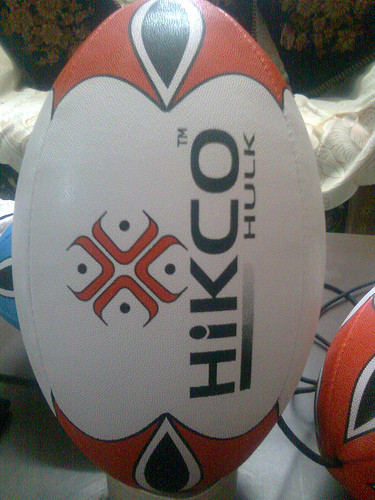} &
\includegraphics[align=c,height=0.18\columnwidth]{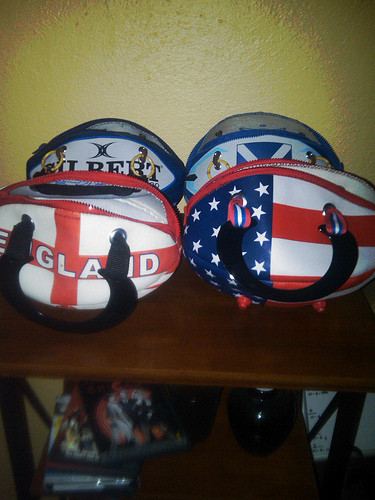} \vspace{0.08cm} \\
\rotatebox[]{90}{\scriptsize Ground truth} &
\includegraphics[align=c,height=0.18\columnwidth]{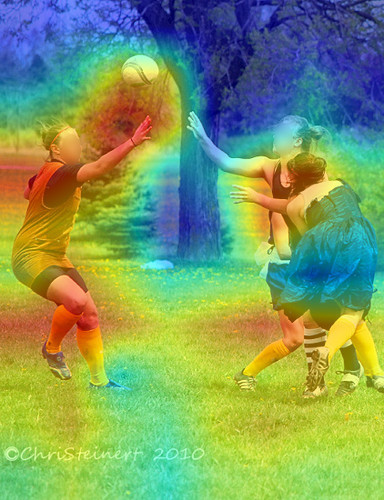} &
\includegraphics[align=c,height=0.18\columnwidth]{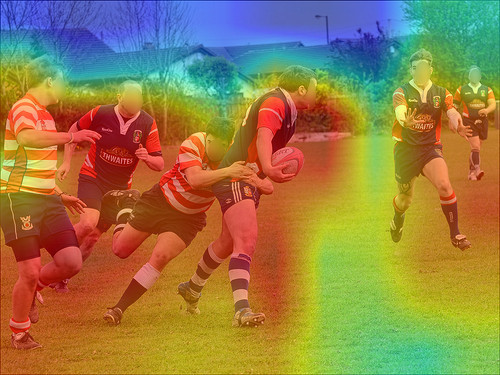} &
\includegraphics[align=c,height=0.18\columnwidth]{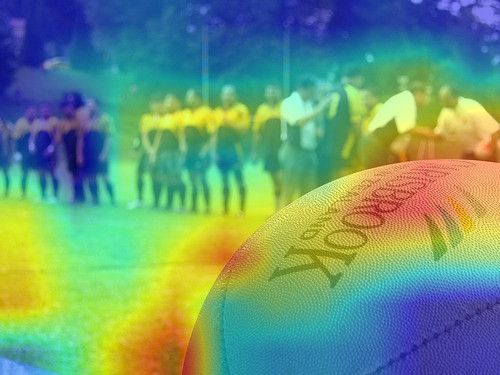} &
\includegraphics[align=c,height=0.18\columnwidth]{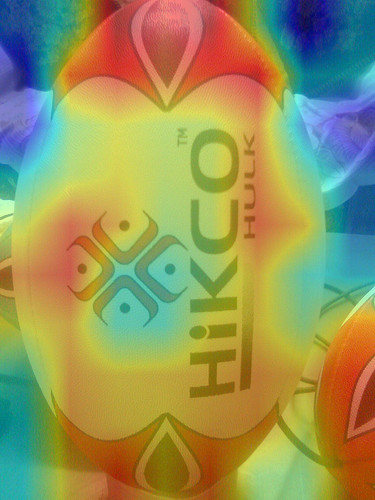} &
\includegraphics[align=c,height=0.18\columnwidth]{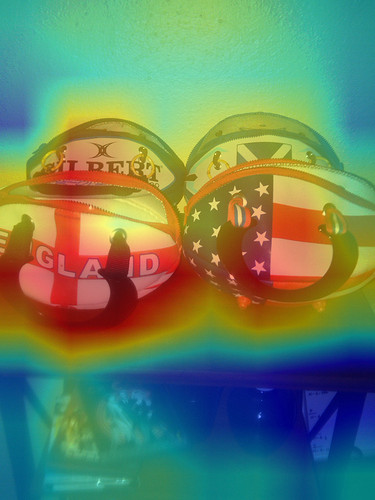} \vspace{0.08cm} \\
\rotatebox[]{90}{{\scriptsize Predicted}} &
\multicolumn{3}{c}{{\scriptsize \textit{Predicted SM = Ground truth SM (``rugby ball'')}}}&
\includegraphics[align=c,height=0.18\columnwidth]{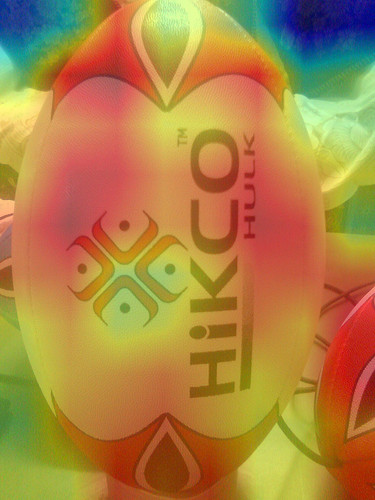} &
\includegraphics[align=c,height=0.18\columnwidth]{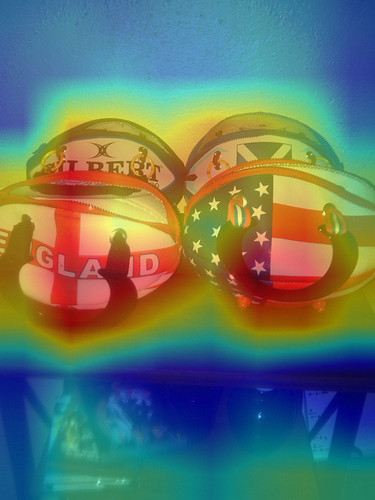} \vspace{0.08cm} \\
&
&
&
&
{\scriptsize bobsled} &
{\scriptsize football helmet}
\end{tabular}
\end{center}
\caption{Images and SMs (superimposed on the input image) from the category ``rugby ball'.
We observe that the classifier mostly learns the environment where the rugby activity takes place,
e.g., football field, rugby players, playing rugby, rather than the rugby ball itself.
In the absence of these clues the classifier fails to categorize correctly the image,
as shown in the examples of the last two columns of the figure.
}
\label{fig:biasRugbyball}
\end{figure}
\begin{figure}[!h]
\begin{center}
\begin{tabular}{cccccc}
\rotatebox[]{90}{\scriptsize Image} &
\includegraphics[align=c,height=0.14\columnwidth]{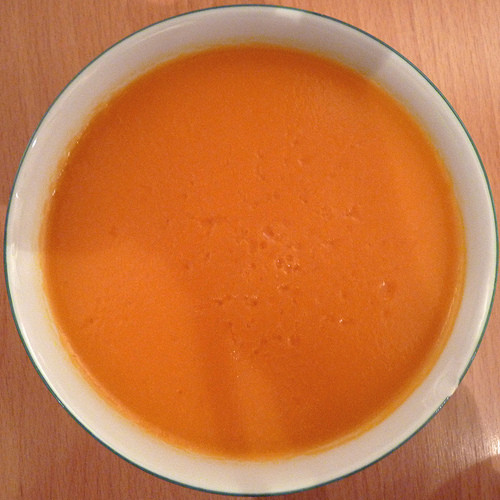} &
\includegraphics[align=c,height=0.14\columnwidth]{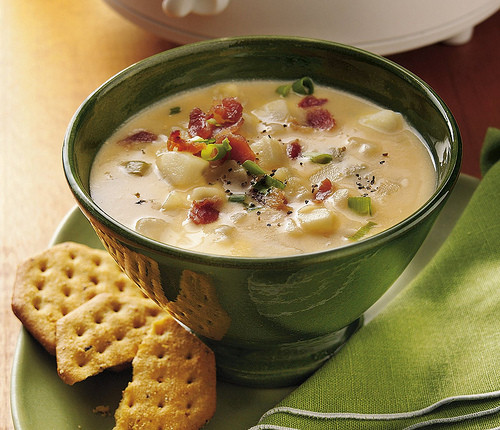} &
\includegraphics[align=c,height=0.14\columnwidth]{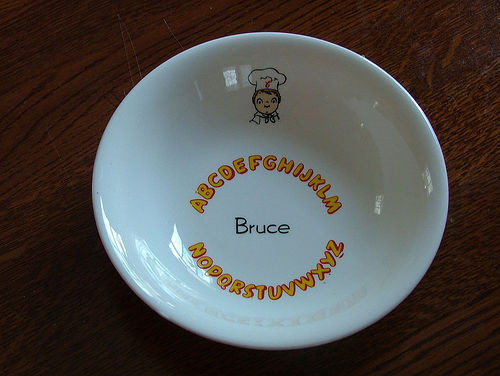} &
\includegraphics[align=c,height=0.14\columnwidth]{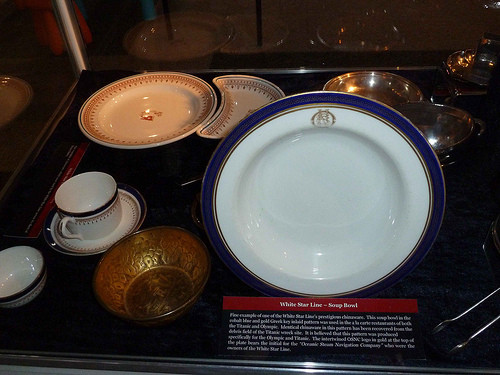} &
\includegraphics[align=c,height=0.14\columnwidth]{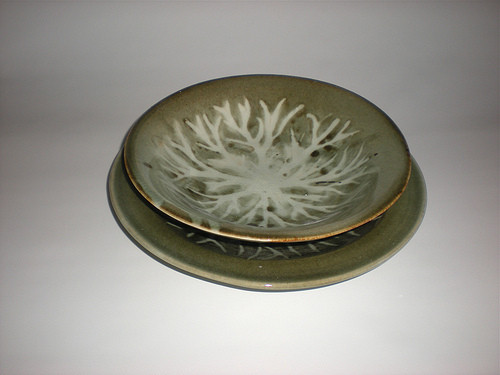} \vspace{0.08cm} \\
\rotatebox[]{90}{\scriptsize Ground truth} &
\includegraphics[align=c,height=0.14\columnwidth]{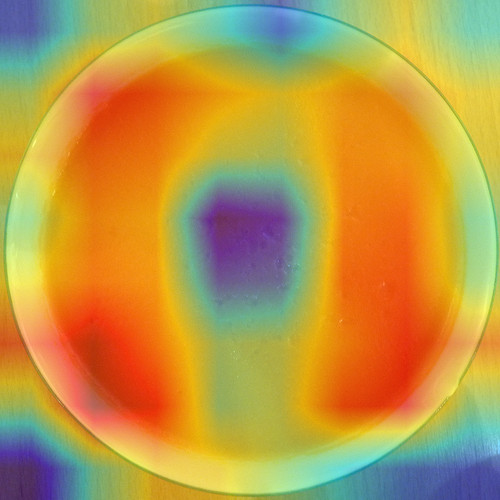} &
\includegraphics[align=c,height=0.14\columnwidth]{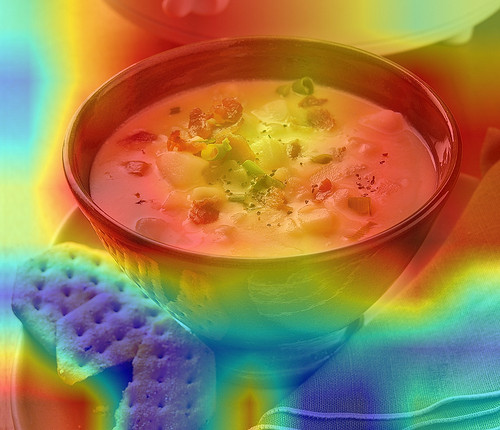} &
\includegraphics[align=c,height=0.14\columnwidth]{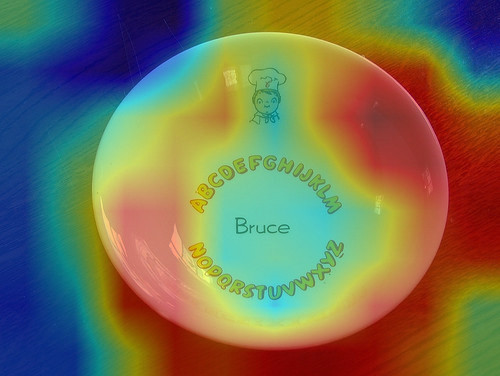} &
\includegraphics[align=c,height=0.14\columnwidth]{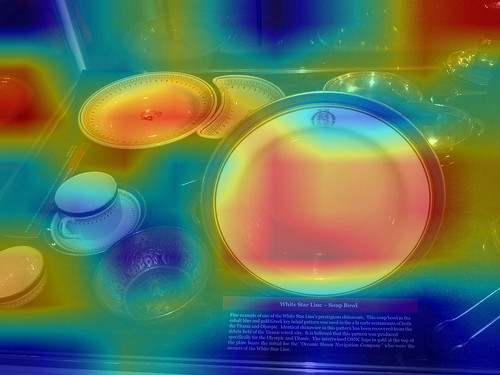} &
\includegraphics[align=c,height=0.14\columnwidth]{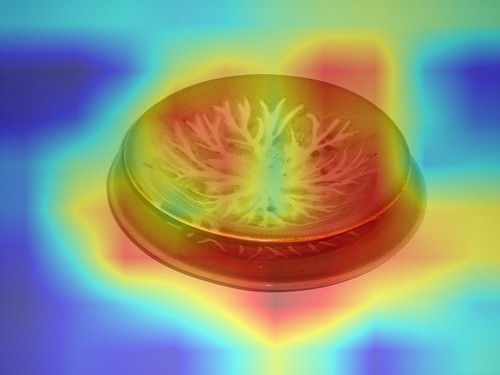} \vspace{0.08cm} \\
\rotatebox[]{90}{{\scriptsize Predicted}} &
\multicolumn{2}{c}{\parbox{2.2cm}{\scriptsize \textit{Predicted SM = Ground truth SM (``soup bowl'')}}}&
\includegraphics[align=c,height=0.14\columnwidth]{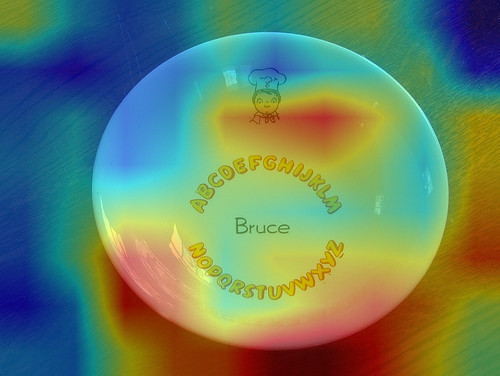} &
\includegraphics[align=c,height=0.14\columnwidth]{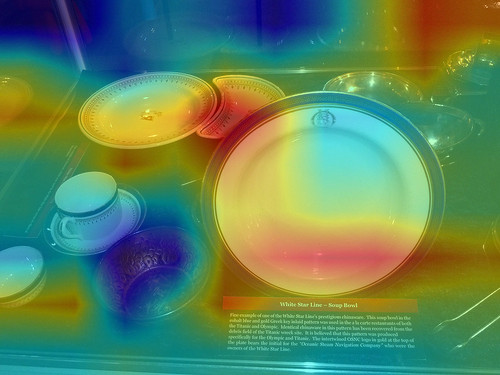} &
\includegraphics[align=c,height=0.14\columnwidth]{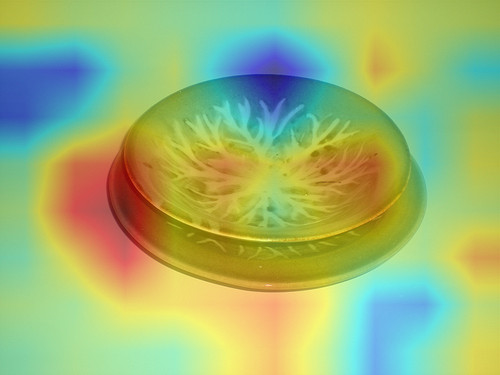} \vspace{0.08cm} \\
&
&
&
{\scriptsize tray} &
{\scriptsize tray} &
{\scriptsize face powder}
\end{tabular}
\end{center}
\caption{Images and SMs (superimposed on the input image) from the category ``soup bowl''. 
We see that the classifier has learned to classify to this category the soup bowls when they contain food;
contrarily, empty soup bowls are miscategorized to other classes such as tray and face powder.
}
\label{fig:biasSoupBowl}
\end{figure}
\begin{figure}[!h]
\begin{center}
\begin{tabular}{cccccc}
\rotatebox[]{90}{\scriptsize Image} &
\includegraphics[align=c,height=0.17\columnwidth]{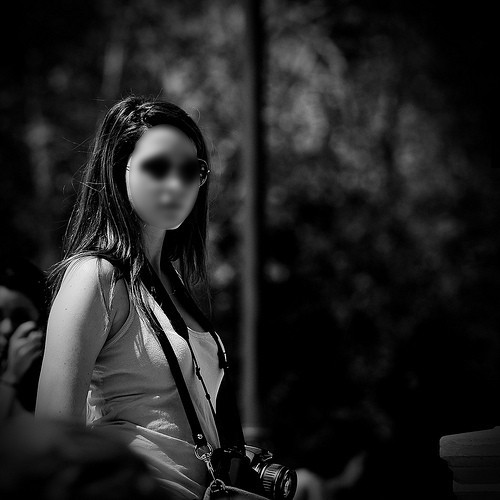} &
\includegraphics[align=c,height=0.17\columnwidth]{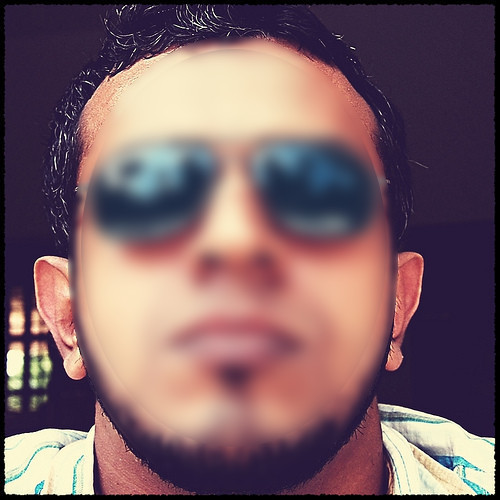} &
\includegraphics[align=c,height=0.17\columnwidth]{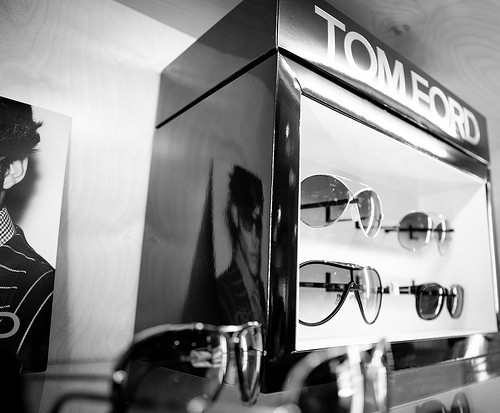} &
\includegraphics[align=c,height=0.17\columnwidth]{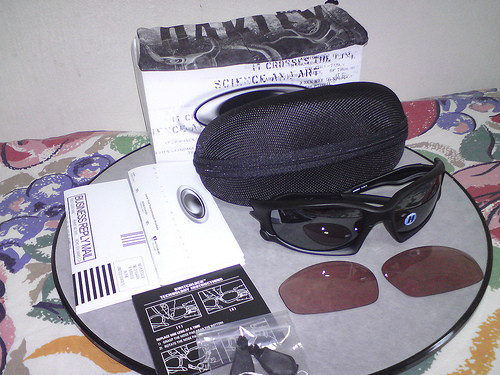} &
\includegraphics[align=c,height=0.17\columnwidth]{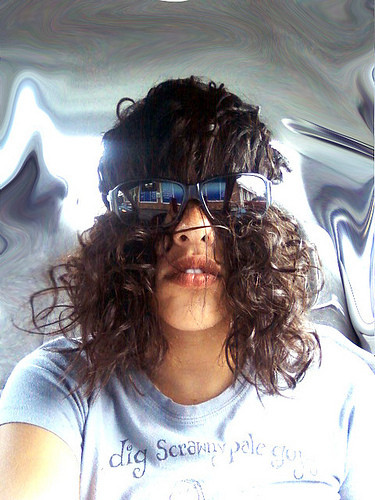} \vspace{0.08cm} \\
\rotatebox[]{90}{\scriptsize Ground truth} &
\includegraphics[align=c,height=0.17\columnwidth]{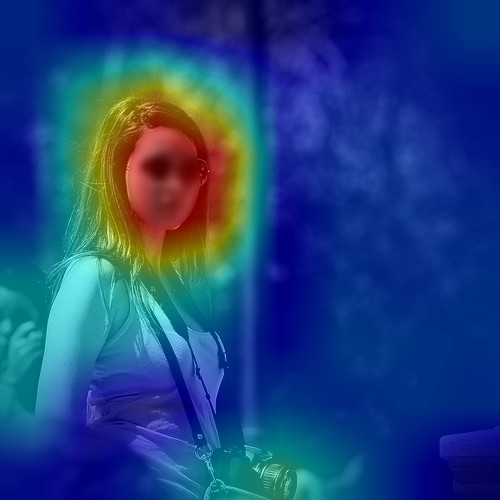} &
\includegraphics[align=c,height=0.17\columnwidth]{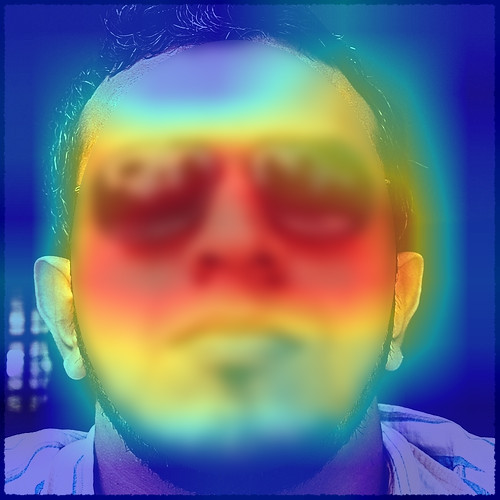} &
\includegraphics[align=c,height=0.17\columnwidth]{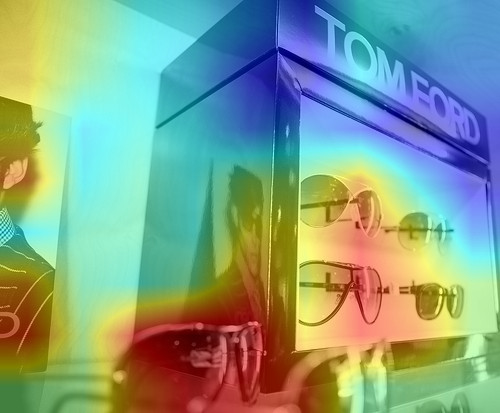} &
\includegraphics[align=c,height=0.17\columnwidth]{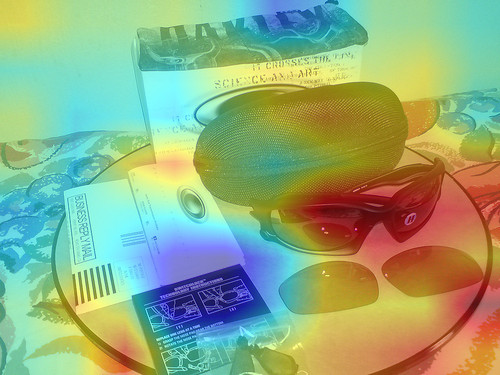} &
\includegraphics[align=c,height=0.17\columnwidth]{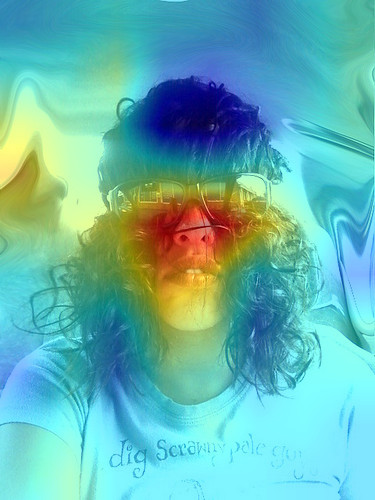} \vspace{0.08cm}\\
\rotatebox[]{90}{{\scriptsize Predicted}} &
\multicolumn{2}{c}{\parbox{2.2cm}{\scriptsize \textit{Predicted SM = Ground truth SM  (``sunglass'')}}}&
\includegraphics[align=c,height=0.17\columnwidth]{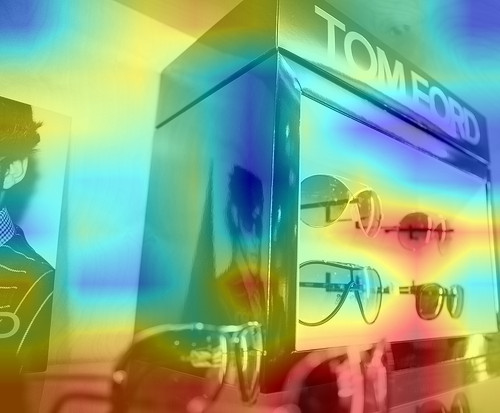} &
\includegraphics[align=c,height=0.17\columnwidth]{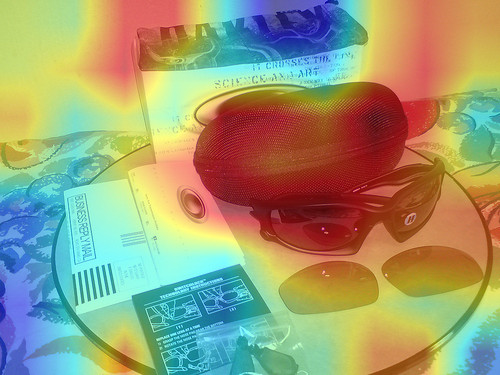} &
\includegraphics[align=c,height=0.17\columnwidth]{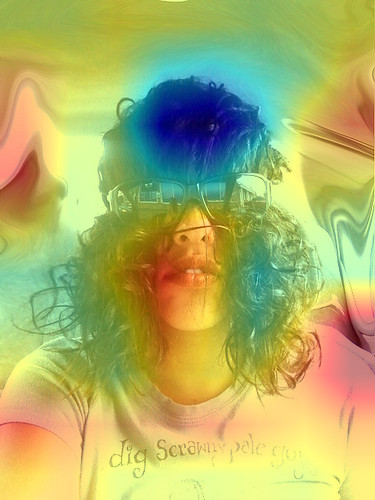} \vspace{0.08cm} \\
&
&
&
{\scriptsize shoe shop} &
{\scriptsize loudspeaker} &
{\scriptsize seatbelt}
\end{tabular}
\end{center}
\caption{Images and SMs (superimposed on the input image) from the category ``sunglass''.
We see that the classifier tends to focus on the sunglasses and the surrounding human face region;
when the relevant human face region does not appear in the image or is occluded,
the classifier infers the wrong category (e.g. shoe shop, loudspeaker).
}
\label{fig:biasSunglass}
\end{figure}

\subsubsection{Comparisons and ablation study:}

The evaluation results in terms of AD($\nu$) and IC($\nu$) for different thresholds
$\nu$ at $\phi_{\nu}()$, i.e., $\nu = 100\%, 50\%$ and $15\%$,
are depicted in the upper and lower half of Table \ref{tab:Results} for VGG-16 and ResNet-50, respectively.
As an ablation study, we also report results for the proposed methods when trained using only
the conventional CE loss, denoted as L-CAM-Fm* and L-CAM-Img*.
The number of forward passes, \#FW, needed to compute the SM
for an input image at the inference stage, is also shown at the last column of this table.
We should note that the auxiliary masks used by RISE (Fig. \ref{fig:XAI_approaches}c) in the VGG-16 experiment
are of size $7 \times 7$ \cite{RISE} (which contrasts to the other approaches that use $14 \times 14$ feature maps for this experiment).
For a fair comparison, we performed an additional experiment with
the $7 \times 7$ feature maps after the last max pooling layer of VGG-16
using our L-CAM-Img, denoted as L-CAM-Img$^\dagger$.
The results of this experiment are reported in the last row of the upper half of Table \ref{tab:Results},
under the L-CAM-Img's results (i.e. the ones obtained using the $14 \times 14$ feature maps).
Moreover, qualitative results for the SMs produced by the different methods
for six sample input images are shown in Fig.~\ref{fig:explanationTwoImg},
while class-specific SMs results for two images containing instances
of two different classes are provided in Fig.~\ref{fig:explanationClassSpecific}.
From the obtained results we observe the following:

i) The proposed L-CAM-Img generally outperforms the gradient-based approaches
and is comparable in AD, IC scores to the perturbation-based approaches Score-CAM, RISE,
yet contrarily to the latter requires only one FW instead of 512-8000 at the inference stage.

ii) L-CAM-Img$^\dagger$ using $7 \times 7$ feature maps achieves the best performance in VGG-16;
our approach is learning-based and, as the experiments showed,
it is easier for it to learn the combination of the feature maps in the lower-dimensional space.
This is consistent with the typical behaviour of learning methods when working with high-dimensional data
that may lay in a low-dimensional manifold (which is often the case with images), i.e. the curse of dimensionality.

iii) L-CAM-Img outperforms L-CAM-Fm, but the latter still generally outperforms the gradient-based approaches.

iv) The proposed approaches provide smooth SMs focusing on important regions of the image, as illustrated in Fig.~\ref{fig:explanationTwoImg}
(and also shown from the very good results obtained for $\nu=15\%$ in Table \ref{tab:Results})
and can produce class-specific explanations, as depicted in the examples of Fig.~\ref{fig:explanationClassSpecific}.

v) From the ablation study of employing only the conventional CE loss (L-CAM-Fm*, L-CAM-Img*),
we conclude that incorporating the two additional terms in the loss function (Eq. (\ref{E:Loss}))
is very beneficial, especially for smaller values of $\nu$.

\subsubsection{Qualitative analysis and discussion:}

In the following, a qualitative study is performed using the proposed L-CAM-Img$^\dagger$.
Specifically, the proposed approach is used to produce visual explanations with $\nu = 100\%$
in order to understand why the VGG-16 classifier may fail to categorize a test image correctly.
To this end, we group the different classification error cases into three categories:

i) \textit{Related classes}: Some Imagenet classes are very close to each other both semantically and/or in appearance. 
For instance, there are classes such as``maillot'' and ``bikini'', ``seacoast'' and ``promontory'',
``schooner'' and ``yawl'',  ``cap'' and ``coffee mug'', and others.
The same is also true for many animals, e.g.,  ``miniature poodle'' and ``toy poodle'',
``panther'' and ``panthera tigris'', ``African elephant'' and ``Indian elephant'', etc.
A few representative examples of images belonging to classes of this category are shown in Fig. \ref{fig:similarClasses}.
In each column of this figure, the input image is presented along with the SM
(superimposed on the input image) corresponding to the ground truth and the predicted labels.
Moreover, under each SM we provide the corresponding class name.
From these examples, an interesting conclusion is that the SMs corresponding to the ground truth and predicted class are similar,
i.e., in both cases the classifier focuses on the same image regions to infer the label of the image.

ii) \textit{Multilabel images}:
Imagenet is a single label dataset, i.e., each image is annotated with only one class label.
However, some images may contain instances belonging to more than one Imagenet class.
For instance, we have identified images containing together instances of the classes ``screw'' and ``screwdriver'',
``warplane'' and ``aircraft carrier'',  ``pier'' and ``boathouse'', and other.
For these cases, the classifier may correctly detect the instance of a class that is visible
but does not correspond to the label of the image, which is considered a classification error.
A few such examples are shown in Fig. \ref{fig:multilabelSamples}.
In contrast to the previous classification error category (related classes),
we observe that now the SMs of the ground truth and predicted class
differ significantly and usually identify a different region of the image.

iii) \textit{Class bias}:
Finally, we performed an analysis of the classification results in order to discover possible biases
on specific Imagenet classes and understand how these biases affect the classifier decisions.
To this end, representative examples from three classes are depicted in Figs. \ref{fig:biasRugbyball}, \ref{fig:biasSoupBowl}, \ref{fig:biasSunglass}.
From Fig. \ref{fig:biasRugbyball} we observe that the classifier
has difficulty inferring the ``soup bowl'' label when an empty soup bowl is depicted.
This finding, together with a visual inspection of the positive training samples for this class,
which reveals that the training set is dominated with images depicting soup bowls filled with food,
indicates that the classifier has in fact learned to detect mostly
the ``soup bowl filled with food'' class instead of the more  general ``soup bowl''.
Similarly, from the results shown in Figs. \ref{fig:biasSoupBowl} and \ref{fig:biasSunglass} we get clear indications that
in place of target class ``rugby ball'' the classifier to a large extent
has learned to detect a broader ``rugby game'' class;
and in place of target class ``sunglass'' the classifier has learned to detect the narrower
``human face wearing sunglasses'' class:
when the sunglasses are shown but the human face is not visible, the classifier fails.

\section{Conclusion} \label{CONCLUSION}

Two new visual XAI methods were presented, which, in contrast to current approaches in this domain,
train an attention mechanism in a supervised manner to produce explanations.
We showed that it is possible to learn the feature maps' weights for deriving very good CAM-based explanations.
We also performed a qualitative study using the explanations produced by our approach
to shed light on the reasons why an image is misclassified,
obtaining interesting conclusions, including the discovery of possible biases in the training set.
As future work we plan to utilize the explanation masks generated by the proposed approaches
for the automatic detection of bias, e.g., extending the work presented in \cite{SernaPMF20}.

\section*{Acknowledgments} \label{Ack}

This work was supported by the EU Horizon 2020 programme under grant agreements H2020-101021866 CRiTERIA and H2020-951911 AI4Media.

\clearpage
%
%


\end{document}